\pdfoutput=1

\documentclass[11pt]{article}

\usepackage{acl}
\usepackage{cognac_v1.0.6}

\usepackage{booktabs}

\usepackage{times}
\usepackage{latexsym}

\usepackage[most]{tcolorbox}

\usepackage[T1]{fontenc}

\usepackage[utf8]{inputenc}

\usepackage{microtype}

\usepackage{inconsolata}

\usepackage{graphicx}

\usepackage{booktabs}
\usepackage{multirow} 

\title{COGNAC at SemEval-2026 Task 5: LLM Ensembles for Human-Level Word Sense Plausibility Rating in Challenging Narratives}

\author{Azwad Anjum Islam \& Tisa Islam Erana\\
        Florida International University \\
        Knight Foundation School of Computing and Information Sciences \\
        11200 SW 8\textsuperscript{th} Street, Miami, FL 33199, USA \\
        \{aisla028, tisla016\}@fiu.edu
}

\begin{document}
\maketitle

\begin{abstract}
We describe our system for SemEval-2026 Task 5, which requires rating the plausibility of given word senses of homonyms in short stories on a 5-point Likert scale. Systems are evaluated by the unweighted average of accuracy (within one standard deviation of mean human judgments) and Spearman Rank Correlation. We explore three prompting strategies using multiple closed-source commercial LLMs: (i) a baseline zero-shot setup, (ii) Chain-of-Thought (CoT) style prompting with structured reasoning, and (iii) a comparative prompting strategy for evaluating candidate word senses simultaneously. Furthermore, to account for the substantial inter-annotator variation present in the gold labels, we propose an ensemble setup by averaging model predictions. Our best official system, comprising an ensemble of LLMs across all three prompting strategies, placed 4$^{th}$ on the competition leaderboard with 0.88 accuracy and 0.83 Spearman's~$\rho$ (0.86 average). Post-competition experiments with additional models further improved this performance to 0.92 accuracy and 0.85 Spearman's~$\rho$ (0.89 average). We find that comparative prompting consistently improved performance across model families, and model ensembling significantly enhanced alignment with mean human judgments, suggesting that LLM ensembles are especially well suited for subjective semantic evaluation tasks involving multiple annotators.
\end{abstract}

\section{Introduction}
\label{sec:intro}

The task of word sense disambiguation (WSD) refers to determining the correct sense of a homonymous word in context. In naturally occurring discourse such as narratives, however, multiple senses of a homonym may be plausible. Human interpretations in such cases often reflect graded preferences rather than a singular objective choice. To address this, \citet{gehring2025ambistory} introduced \textit{AmbiStory}, a dataset in which a target sentence is constructed to be ambiguous while surrounding sentences provide indirect cues that shift the relative plausibility of competing senses. Annotators then assigned plausibility ratings on a 1--5 Likert scale, allowing multiple senses to receive non-zero support. SemEval-2026 Task 5 \cite{janosch2026Semeval} builds on this dataset. Given a story context with a target homonym and a candidate sense, systems must predict plausibility ratings on the same 5-point scale as human annotators: \(1\) (implausible), \(2\) (less plausible than other senses), \(3\) (one of several equally plausible senses), \(4\) (more plausible than other senses), and \(5\) (only plausible sense). Evaluation uses the average of two metrics: accuracy within one standard deviation of the mean human judgment, and Spearman correlation.

We explore three prompting strategies using six closed-source LLMs for this task: (i) a zero-shot baseline, (ii) Chain-of-Thought (CoT) prompting with structured intermediate reasoning, and (iii) comparative prompting that jointly scores competing senses using a single prompt. To address the significant annotation variation in the gold human ratings, we further propose an LLM ensemble that aggregates predictions from different models and prompting strategies via unweighted average. Our best official submission ranked 4$^{th}$ with an average score of 0.86 (0.88 accuracy, 0.83 Spearman’s $\rho$); adding four additional models post-competition improved this to 0.89 (0.92 accuracy, 0.85 Spearman’s $\rho$). On the development set, we find that comparative prompting consistently outperformed the zero-shot baseline and CoT strategies. Finally, ensembles produced superior performance to individual models in general, with ensembles of small models often aligning better with aggregated human judgments than larger, more capable models.

Our contributions are threefold. First, we evaluate three prompting strategies using a total of ten LLMs across three model families for predicting graded sense plausibility in narrative contexts. Second, we show that evaluating competing senses jointly in a single comparative prompt generally improves performance relative to a zero-shot baseline, while Chain-of-Thought prompting does not. Third, we also demonstrate that simple LLM ensembling significantly improves alignment with aggregated human judgments in a task characterized by substantial annotation variation.

The remainder of this paper is structured as follows. Section~\ref{sec:relatedwork} first provides a background on word sense disambiguation and LLM-based approaches. Section~\ref{sec:task} describes the task and dataset. Section~\ref{sec:approach} presents our prompting strategies and ensemble method. Section~\ref{sec:results} reports development and test results, including post-competition improvements. Finally, Sections~\ref{sec:discussion} and~\ref{sec:limitation} discuss findings and limitations.

\section{Related Work}
\label{sec:relatedwork}
Early work on word sense disambiguation (WSD) treated sense interpretation as single-label classification over lexical resources such as WordNet~\cite{miller1995wordnet}, with systems evaluated on recovering one correct sense per token in context~\cite{Navigli2009Wordnet, bevilacqua2021recent}. Parallel work on graded word sense annotation and WSD evaluation instead allowed annotators to distribute applicability over multiple senses, challenging the assumption of a single best-fitting sense and providing evaluation schemes that directly target graded plausibility~\cite{erk2009graded,jurgens2012graded}. More recent WSD methods relied on contextualized encoders such as BERT and gloss-aware architectures, but still typically assumed discrete sense decisions rather than modeling graded plausibility across competing readings~\cite{hadiwinoto2019glossbert, loureiro2021analysis}. Recent studies have also evaluated prompting-based and instruction-tuned large language models directly on standard WSD benchmarks, showing that carefully designed prompts can match or approach the performance of dedicated WSD systems while maintaining greater flexibility in how senses are represented~\cite{sumanathilaka2024gptwsd,meconi2025llmwsd}. \textit{AmbiStory}~\cite{gehring2025ambistory} extended WSD to narrative contexts where multiple senses can remain simultaneously plausible and human ratings exhibit substantial variation even within a single story. In contrast to earlier narrative benchmarks~\cite{mostafazadeh-etal-2016-corpus}, AmbiStory considered under-specification and disagreement as central phenomena rather than noise, aligning with broader work that treats ambiguity and disagreement in semantic annotation as informative signals rather than annotation error~\cite{pavlick2019inherent}. 

Prompting-based LLM methods offer a flexible alternative to WSD-specific architectures. Chain-of-Thought (CoT) prompting in particular has been shown to improve multi-step reasoning by eliciting intermediate rationales across multiple benchmarks~\cite{wei2022chain_of_thought}. However, prior CoT work commonly assumed objective ground-truth labels, leaving it unclear how structured reasoning interacts with subjective, graded judgments such as those in AmbiStory. Meanwhile, a growing body of work on LLM-as-a-judge frameworks has explored pairwise or list-wise comparisons as alternatives to absolute scoring for evaluating open-ended outputs~\cite{li2025llmjudge, lu2024ahp, zhou2024fairer}. This line of work suggests that relative preference judgments are often more reliable than direct scoring, motivating comparative prompting schemes in which models assess competing interpretations to make more informed judgments. Finally, ensemble-style approaches and sampling-based strategies aggregate multiple LLM outputs to reduce variance and improve reliability over single-pass decoding~\cite{dietterich2000ensemble, wang2023self_consistency}. While these methods implicitly approximate distributions over reasoning paths, they have rarely been applied to explicitly model human disagreement over linguistic interpretations, despite growing evidence that such disagreement persists even under carefully controlled conditions~\cite{pavlick2019inherent}. 

\begin{table*}[t]
\small
\centering
\renewcommand{\arraystretch}{1.1}
\begin{tabular*}{\textwidth}{@{\extracolsep{\fill}} l c c c  c c c | c c c}
\toprule
\textbf{Model} 
  & \multicolumn{3}{c}{\textbf{Accuracy}} 
  & \multicolumn{3}{c}{\textbf{Spearman $\rho$}} 
  & \multicolumn{3}{c}{\textbf{Average Score}} \\
\cmidrule(lr){2-4} \cmidrule(lr){5-7} \cmidrule(lr){8-10}
 & Zero-shot & CoT & Comp. & Zero-shot & CoT & Comp. & Zero-shot & CoT & Comp. \\
\midrule
gpt-5-nano            & .72 & .68 & .82    & .76 & .66 & .77    & .74 & .67~$\downarrow$ & .80~$\uparrow$  \\
gpt-4.1-mini          & .72 & .73 & .74    & .68 & .75 & .74    & .70 & .74~$\uparrow$   & .74~$\uparrow$  \\
gemini-2.5-flash      & .69 & .62 & .71    & .75 & .73 & .76    & .72 & .67~$\downarrow$ & .73~$\uparrow$  \\
gpt-4o-mini           & .72 & .61 & .75    & .70 & .62 & .69    & .71 & .61~$\downarrow$ & .72~$\uparrow$  \\
gemini-2.0-flash      & .59 & .61 & .68    & .64 & .68 & .70    & .61 & .65~$\uparrow$   & .69~$\uparrow$  \\
gemini-2.5-flash-lite & .65 & .53 & .64    & .59 & .55 & .65    & .62 & .54~$\downarrow$ & .64~$\uparrow$  \\
\midrule
gpt-5-mini$^*$        & .83 & .71 & .80    & .80 & .79 & .80    & .81 & .75~$\downarrow$ & .80~$\downarrow$ \\
Deepseek-v3.2$^*$     & .77 & .81 & .82    & .76 & .76 & .75    & .76 & .78~$\uparrow$   & .79~$\uparrow$  \\
gpt-5.1$^*$           & .76 & .73 & .78    & .78 & .79 & .78    & .77 & .76~$\downarrow$ & .78~$\uparrow$  \\
gpt-4o$^*$            & .74 & .71 & .77    & .71 & .77 & .74    & .72 & .74~$\uparrow$   & .75~$\uparrow$  \\
\midrule
Average (All Models) & 0.72 & 0.67 & \textbf{0.75} & 0.72 & 0.71 & \textbf{0.74} & 0.72 & 0.69~$\downarrow$ & \textbf{0.74}~$\uparrow$ \\
\bottomrule

\end{tabular*}
\caption{Individual model performance under different prompting strategies on the development set. Arrows ($\uparrow$, $\downarrow$) indicate changes relative to the zero-shot prompting strategy. The horizontal divider separates models used only in post-competition experiments, marked with (*).}
\label{tab:model-results}
\vspace{-0.5em}
\end{table*}

\section{SemEval-2026 Task 5}
\label{sec:task}

The task uses the AmbiStory dataset~\cite{gehring2025ambistory}, comprising 3,798 samples across 633 setups, where each setup features a target homonym embedded in an ambiguous short story consisting of either four (open-ended) or five (ended) sentences. For each setup, there are three story variants: one open-ended version without an ending sentence and two ended versions, each tailored to one of the two candidate senses. Combining the three ending variants with the two candidate senses yields exactly six samples  $(3 \times 2 = 6)$ per setup and 3,798 samples in total. Word sense pairs where one sense involves a different part of speech than the other or requires a specific particle were removed.

Gold plausibility ratings are averages of 5--6 human annotations per sample, where annotators rate word senses on a 5-point Likert scale given the setup. Inter-annotator agreement, measured using Krippendorff’s $\alpha$, is 0.506, and the average standard deviation per sample is $\sigma = 0.946$. The dataset is split into a labeled training set (2,280 samples), a development set (588 samples), and an unlabeled test set (930 samples). Test set labels were made available post-competition. Notably, homonyms do not appear across multiple splits,  i.e, a homonym that appears in test data, for example, does not appear in training or development set. 

Given a target homonym, story context, and candidate sense, systems must predict a plausibility score between 1 and 5. Performance is measured by the unweighted average of two metrics: (i) Accuracy within Standard Deviation, i.e., the proportion of predictions within one standard deviation of the mean human ratings, and (ii) Spearman correlation between predictions and mean human ratings.

\section{Approach}
\label{sec:approach}
We evaluated three prompting strategies together with an ensemble method that averages predictions across models and prompting strategies. In our initial experiments, we used three GPT models and three Gemini models\footnote{\texttt{gpt-4o-mini}, \texttt{gpt-4.1-mini}, \texttt{gpt-5-nano}, \texttt{gemini-2.0-flash}, \texttt{gemini-2.5-flash}, and \texttt{gemini-2.5-flash-preview}.}. Following the competition, we extended our experiments with four additional models\footnote{\texttt{gpt-4o}, \texttt{gpt-5-mini}, \texttt{gpt-5.1}, and \texttt{deepseek-v3.2}.}. During system development, we relied exclusively on the development set and did not use the larger training set due to cost and time constraints.

\subsection{Prompting Strategies}
\label{sec:prompting}

\textbf{Zero-shot prompting:} We first established a zero-shot prompting setup as a baseline that closely follows the prediction prompt described by \citet{gehring2025ambistory} with few modifications. Notably, we did not include few-shot examples in the prompt (Appendix~\ref{prompt:baseline}). The authors report mixed effects of few-shot examples and note that they may even degrade performance for some models (e.g., \texttt{gpt-4o} and \texttt{gpt-4o-mini}). Omitting examples also reduced context length and inference cost, which is particularly beneficial in our multi-model evaluation and ensembling setting. We also required models to output a brief justification for their plausibility judgments in a structured JSON format. We used all models at their default temperature values.


\textbf{Chain-of-Thought prompting:} We adopted a structured Chain-of-Thought (CoT) prompting strategy that decomposes the plausibility judgment into five explicit reasoning steps (Appendix~\ref{prompt:cot}) with the goal of performing a step-by-step evaluation of word senses. We guided the model to (1) characterize the overall context type (e.g., open-ended vs.\ ended, domain/topic), (2) analyze the grammatical and syntactic role of the target word, (3) assess semantic fit between the proposed sense and the setup and ending (if present) of the story, (4) enumerate plausible alternative senses and contrast them against the candidate meaning, and (5) assign a final rating on the plausibility scale of 1--5. We designed this intermediate reasoning framework to reduce ``first-interpretation'' bias by requiring explicit reflection on competing senses, even when one interpretation initially appears obvious. However, this approach did not surpass baseline performance on the development set for six of the ten models (Table~\ref{tab:model-results}). The task’s subjective, human-perceived plausibility judgments likely limit the benefits of CoT, as step-by-step reasoning can steer models toward overly analytical interpretations that diverge from human intuitions.


\begin{figure}[t]
\includegraphics[width=\columnwidth]{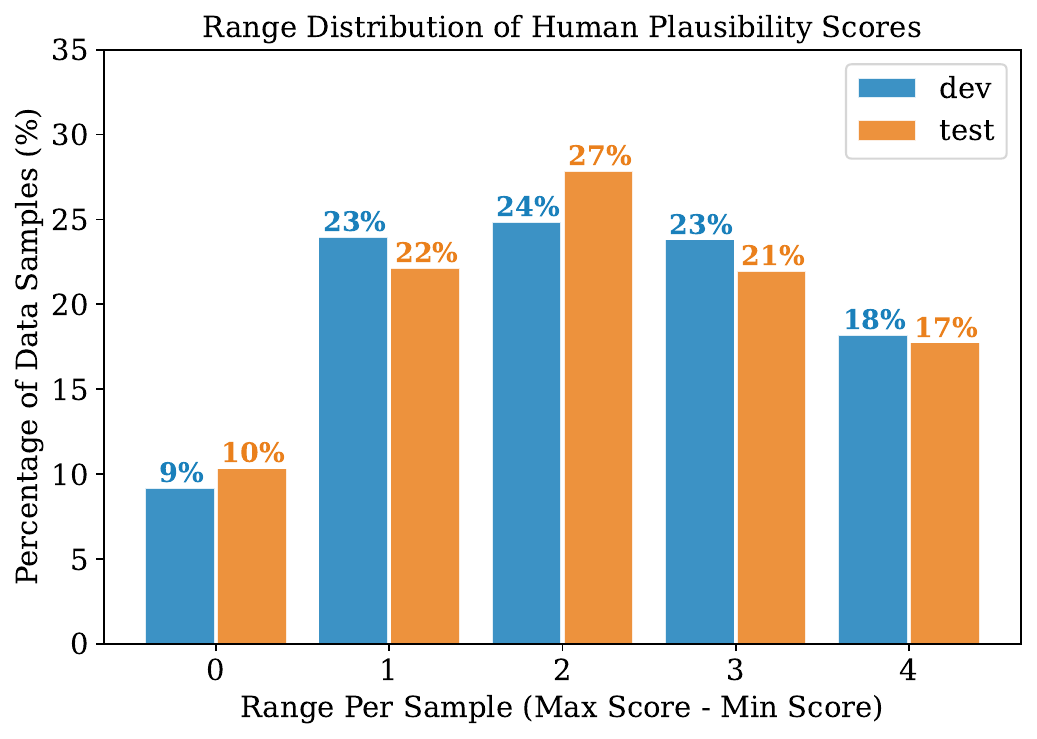}
\caption{Distribution of word sense plausibility scores by human in development and test data in percentage}
\label{fig:range}
\vspace{-0.5em}
\end{figure}

\textbf{Comparative prompting:} \textit{AmbiStory} annotates word senses on a 5-point Likert scale where score definitions are comparative in nature (e.g., ``less plausible than other meanings'', ``one of many similarly plausible meanings''). Moreover, as discussed in Section~\ref{sec:task}, there are two different narrative endings (excluding the open-ended variant) for each story setup where each ending is deliberately constructed to increases the plausibility of one of the two candidate senses of the target homonym. This makes comparative judgment intrinsic to the task as plausibility is defined relative to competing interpretations and contexts that bias one sense over the other. Evaluating a single sense in isolation risks overlooking this insight, as the model may not explicitly consider the other competing sense of the word. Therefore, we adopted a comparative prompting strategy (Appendix~\ref{prompt:comparative}), where we presented both candidate senses together as two options to the model for each (homonym, context, ending) combination. We then asked the model to assign a plausibility score (1--5) to each option, along with a brief justifications. This design encourages explicit comparison between competing senses and aligns model inference with the comparative nature of the annotation scheme. This strategy yields the strongest overall performance on the development set compared to the baseline and CoT strategies (Table~\ref{tab:model-results}). These results are further discussed in Section~\ref{sec:results}.

\begin{table}[t]
\small
\centering
\begin{tabular}{lccc}
\toprule
\textbf{Ensembles} & \textbf{Accuracy} & \textbf{$\rho$} & \textbf{Average} \\
\midrule
$E_{zeroshot}$  &   0.86 & 0.79 & 0.83 \\
$E_{CoT}$       &   0.80 & 0.78 & 0.79 \\
$E_{comp}$      &   0.87 & 0.81 & 0.84 \\
$E_{all}$       &   0.89 & 0.84 & 0.87 \\
\midrule
\multicolumn{4}{l}{Including post-competition models:}\\
$E_{zeroshot}$  &   0.91 & 0.84 & 0.88 \\
$E_{CoT}$       &   0.86 & 0.84 & 0.85 \\
$E_{comp}$      &   0.89 & 0.83 & 0.86 \\
$E_{all}$       &   0.93 & 0.86 & 0.89 \\
\midrule
\multicolumn{4}{l}{Best individual model (for reference):}\\
\texttt{gpt-5-mini}$_{(zeroshot)}$  
                &   0.83 & 0.80 & 0.81 \\
\bottomrule
\end{tabular}
\caption{Performance summary of different LLM ensembles on development data.}
\label{tab:ensemble}
\vspace{-0.5em}
\end{table}

\subsection{LLM Ensemble}
\label{sec:ensemble}
A key challenge of this task is the substantial annotation variability in the gold human judgments. \citet{gehring2025ambistory} reports a relatively low inter-annotator agreement (Krippendorff’s $\alpha=0.506$) while the average per-sample standard deviation ($\sigma=0.946$) is high for a 5-point scale. This is further reflected in the distribution of per-sample range (max rating - min rating) in gold annotations (Figure~\ref{fig:range}), showing that roughly 65\% of samples across development and test data have a range $\ge$2. Figure~\ref{fig:range} also shows that in 17-18\% of samples, human annotators assigned both 1 \textit{(The meaning is not possible at all)} and 5 \textit{(The meaning is the only plausible meaning)} ratings to the same sense of a homonym in the same story setup. The official ``accuracy'' metric partially acknowledges this variability by crediting predictions that fall within one standard deviation of the mean human rating. However, this tolerance does not eliminate the ambiguity completely. For example, if the human ratings are (1,1,4,4,5; range$=$4) or (2,3,3,4,5; range$=$3), then the 1-standard-deviation tolerance excludes ratings such as 1, 2, and 5, even though each is supported by at least one human annotator.

Based on these observations, we propose that a single model may struggle to reproduce mean human judgment in a multi-annotator setup, especially when the judgments are highly subjective. We therefore experiment with aggregating model predictions across multiple LLMs via unweighted average. We construct three ensembles $E_{zeroshot}$, $E_{CoT}$, and $E_{comp}$ by aggregating all model predictions within individual prompting strategies  (\S~\ref{sec:prompting}). In addition, we construct a fourth ensemble $E_{all}$ that aggregates predictions of all models across all strategies.


\begin{table}[t]
\small
\centering
\begin{tabular}{l ccc}
\toprule
\textbf{Team} & \textbf{Accuracy} & \textbf{$\rho$} & \textbf{Avg.} \\
\midrule
SRCB (1$^{st}$)                         & 0.93 & 0.86 & 0.89 \\
\textbf{COGNAC} (4$^{th}$)              & 0.88 & 0.83 & 0.86 \\
\midrule
$E_{all}$ \textit{(post-competition)}   & 0.92 & 0.85 & 0.89 \\
\bottomrule
\end{tabular}
\caption{Our (COGNAC) performance on test data: official submission and post-competition improvements.}
\label{tab:test-results}
\vspace{-0.5em}
\end{table}

\section{Results}
\label{sec:results}

\subsection{Individual LLM Performance}
Our development set experiments are summarized in Table~\ref{tab:model-results}. Comparative prompting outperformed both zero-shot and CoT strategies across all six models during competition, with \texttt{gpt-5-nano} achieving the highest average score of 0.80 (0.82 accuracy, 0.77 $\rho$). Post-competition experiments with four additional models further improved single-model performance. While comparative prompting remained the better strategy for 9 out of 10 models, the best individual result across all 30 (10 models$\times$3 strategies) combinations was 0.81 (0.83 accuracy, 0.80 $\rho$) from \texttt{gpt-5-mini} under the baseline prompt. We also noticed that CoT performed significantly better in the larger reasoning models (e.g., \texttt{deepseek-v3.2}, \texttt{gpt-5.1}).

\subsection{Ensemble Performance}
Table~\ref{tab:ensemble} summarizes ensemble performance on the development set. The ensemble $E_{all}$, which averages predictions across all models and strategies, achieved the best score of 0.87 (0.89 accuracy, 0.84 $\rho$). Post-competition models further improved the average score of $E_{all}$ to 0.89 (0.93 accuracy, 0.86 $\rho$). Although $E_{CoT}$ remained the weakest ensemble, it showed significant improvement with the inclusion of larger reasoning models. 

\begin{figure}[t]
\includegraphics[width=\columnwidth]{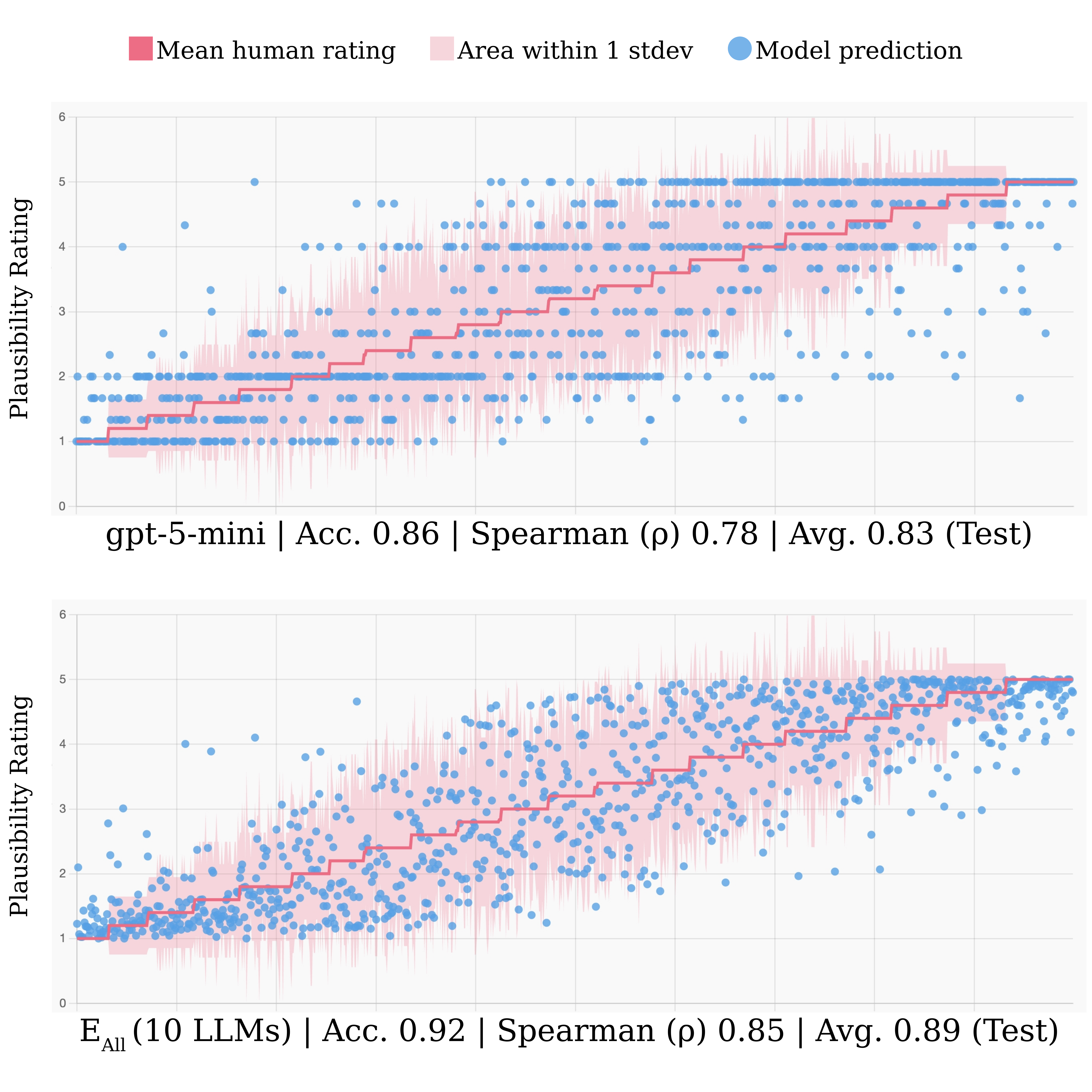}
\caption{Model predictions vs. human ratings (mean $\pm$ 1 stdev) on test data. $E_{all}$ (10 LLMs) aligns better than the best individual model (\texttt{gpt-5-mini}).}
\label{fig:ensemble-compare}
\vspace{-0.5em}
\end{figure}

Figure~\ref{fig:ensemble-compare} shows how $E_{all}$ conforms to the human judgment area (mean $\pm$ 1-stdev) much better than even the best individual model \texttt{gpt-5-mini}. Notably, we find ensemble of even the three weakest models in our experiments (\texttt{gemini-2.0-flash}, \texttt{gemini-2.5-flash-lite}, \texttt{gpt-4o-mini}) still reached an average score of 0.812 on the development data, rivaling the best single-model performance. We release an interactive dashboard\footnote{\url{https://otpid.github.io/cognac-semeval-2026-5-dash/}} with all possible model combinations on both development and test sets to facilitate further exploration of ensemble performance. 

We submitted the ensemble $E_{all}$ with the initial six LLMs to the competition, which placed 4$^{th}$ on the leaderboard. Performance of this submission along with post-competition improvements on the test data are shown in Table~\ref{tab:test-results}. 

\section{Discussion and Conclusion}
\label{sec:discussion}

We analyze LLM capabilities for predicting human plausibility ratings of homonymous words in challenging narrative contexts using ten LLMs and three prompting strategies. Comparative prompting, which jointly evaluates competing senses, consistently outperforms zero-shot baselines across model families, while CoT provides limited gains. Our most noteworthy contribution is demonstrating the effectiveness of model ensembling for predicting mean human judgments in high-variance, multi-annotator settings, with even small-model combinations rivaling top individual performances. Our official submission ranked 4$^{th}$ with 0.86 average score on the leaderboard, with post-competition scaling improving results to 0.89, which is on par with the top leaderboard position. We also make available our results via a public dashboard that enables further exploration of ensemble performances.

\section{Limitations}
\label{sec:limitation}
Our system relies entirely on closed-source commercial LLMs, which limits reproducibility and accessibility due to API costs. These constraints also affected our experimental design, preventing us from using the larger training dataset. In addition, we did not explore parameter fine-tuning despite the availability of labeled data. Finally, our system relies on ensembling across multiple models and prompting strategies, which increases inference-time compute, latency, and operational cost, making the system less practical in resource-constrained settings.

\section{Acknowledgments}
\label{sec:acknowledgments}
We thank Mark A. Finlayson for his invaluable guidance during the development of this project and the preparation of this paper.

\bibliography{paper}

\clearpage
\appendix

\section{Zero-shot Prompt Template}
\label{prompt:baseline}
Prompt template used in the zero-shot prompting strategy.
\begin{tcolorbox}[ breakable, colback=white, colframe=black, boxrule=0.5pt, arc=0pt ]
\ttfamily
Determine how plausible a meaning of a word is in the context of a short text. 
You will see a short text in which one sentence is marked with "**". That sentence contains a word that can typically take on multiple different meanings, depending on the context. You will be given one such meaning.\\

Your task is to determine a score between 1-5 judging the plausibility of the given meaning:\\
1 (The given meaning is not plausible at all given the context.)

2 (The given meaning is theoretically conceivable, but less plausible than other meanings.)

3 (The given meaning represents one of multiple, similarly plausible interpretations.)

4 (The given meaning represents the most plausible interpretation; other meanings may still be conceivable.)

5 (The given meaning is the only plausible meaning given the context.) \\

Provide a short reasoning behind your score. Output should be a valid JSON string like this:\\

\{
    "answer" : (number between 1 to 5),\\
    "reason" : (a short explanation)
\}
\end{tcolorbox}

\section{Chain-of-Thought Prompt Template}
\label{prompt:cot}

Prompt template used in the Chain-of-Thought prompting strategy.
\begin{tcolorbox}[ breakable, colback=white, colframe=black, boxrule=0.5pt, arc=0pt ]
\ttfamily
You are an expert at understanding word meanings in context. Evaluate the plausibility of a word meaning in context.\\

=== STORY ===

\{precontext + context + ending\}

=== EVALUATION ===

Word: \{word\}

Proposed Meaning: \{definition\}. Example: \{example\}\\

**Rating Scale:**\\
1 = Not plausible at all given the context \\
2 = Conceivable but much less plausible than other meanings  \\
3 = One of multiple equally plausible interpretations \\
4 = The most plausible interpretation; other meanings still conceivable \\
5 = The only plausible meaning given the context \\

=== REASONING FRAMEWORK ===

Step 1 - Identify Context Type: \\
- Is this story open-ended (no clear resolution) or ended (clear resolution)? \\
- What domain/topic is the story about? \\

Step 2 - Analyze Word Usage: \\
- How does the sentence structure constrain the word's meaning? \\
- What grammatical role does "\{word\}" play? \\

Step 3 - Evaluate Semantic Fit: \\
- Does this meaning align with the precontext setup? \\
- If there's an ending, does it support or contradict this meaning? \\
- Rate semantic fit: Poor / Moderate / Good / Excellent \\

Step 4 - Consider Alternatives: \\
- What other meanings of "\{word\}" could apply here? \\
- How does the proposed meaning compare to the most likely alternative? \\

Step 5 - Final Rating: \\
Based on the above analysis, rate on 1-5 scale. \\

Format your response in JSON:\\

\{
  "context": [1 sentence],\\
  "semantic\_fit": [Poor/Moderate/Good/Excellent],\\
  "alternatives": [brief note on other possible meanings],\\
  "rating": [1-5]
\}
\end{tcolorbox}

\section{Comparative Prompt Template}
\label{prompt:comparative}

Prompt template used in the comparative prompting strategy.
\begin{tcolorbox}[ breakable, colback=white, colframe=black, boxrule=0.5pt, arc=0pt ]
\ttfamily
A word can have multiple meanings. For example: the word "\{word\}" can have the following two meanings: \\

Option a: \{Meaning 1\}, as in the sentence: \{Example 1\}

Option b: \{Meaning 2\}, as in the sentence: \{Example 2\} \\

In the sentence "\{sentence\}", in the following paragraph: 

\{precontext + sentence + ending\} \\

Judge the plausibility of both options in a score from 1-5:

1: (The meaning is not plausible at all given the context.)

2: (The meaning is theoretically conceivable, but less plausible than other meanings.)

3: (The meaning represents one of multiple, similarly plausible interpretations.)

4: (The meaning represents the most plausible interpretation; other meanings may still be conceivable.)

5: (The meaning is the only plausible meaning given the context.) \\

Your output should be in JSON format, with a score for both options, and reasoning behind your choice of score.

Output format should be: \\

\{
  "a\_score": <1/2/3/4/5>,\\
  "a\_score\_reason": "...",\\
  "b\_score": <1/2/3/4/5>,\\
  "b\_score\_reason": "..."\\
\}
\end{tcolorbox}


\end{document}